\documentclass{vgtc}                          

\graphicspath{{figures/}{pictures/}{images/}{./}} 

\usepackage{times}                     

\usepackage{tabu}                      
\usepackage{booktabs}                  
\usepackage{lipsum}                    
\usepackage{mwe}                       
\usepackage{amsmath}
\usepackage{mathptmx}                  

\onlineid{1149}

\vgtccategory{Research}

\vgtcinsertpkg

\title{UniHands: Unifying Various Wild-Collected Keypoints for Personalized Hand Reconstruction}

\author{Menghe Zhang\thanks{e-mail: menghe.z@samsung.com}\\ %
        \scriptsize Samsung Semiconductor Inc. %
\and Joonyeoup Kim\thanks{e-mail: kim3649@purdue.edu}\\ %
     \scriptsize Samsung Semiconductor Inc. %
\and Yangwen Liang\thanks{e-mail: liang.yw@samsung.com}\\ %
     \scriptsize Samsung Semiconductor Inc. %
\and Shuangquan Wang\thanks{e-mail: shuangquan.w@samsung.com}\\ %
     \scriptsize Samsung Semiconductor Inc. %
\and Kee-Bong Song\thanks{e-mail: keebong.s@samsung.com}\\ %
     \scriptsize Samsung Semiconductor Inc. %
     }

\teaser{
  \centering
  \includegraphics[width=\textwidth]{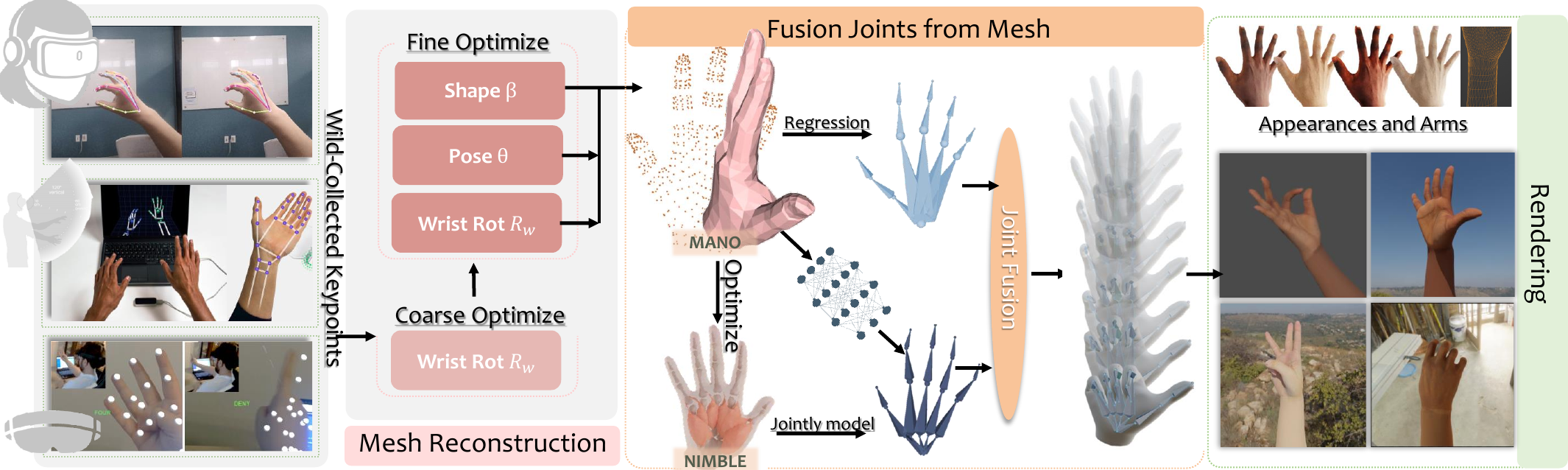}
  \caption{From wild-collected hand keypoints across various sources, UniHands reconstructs personalized, parameterized hand models with unified joints, transforming keypoint-based captures into high-fidelity representations suitable for diverse hand-related applications.}
  \label{fig:teaser}
}

\abstract{
Accurate hand motion capture and standardized 3D representation are essential for various hand-related tasks. Collecting keypoints-only data, while efficient and cost-effective, results in low-fidelity representations and lacks surface information. Furthermore, data inconsistencies across sources challenge their integration and use.
We present UniHands, a novel method for creating standardized yet personalized hand models from wild-collected keypoints from diverse sources. Unlike existing neural implicit representation methods, UniHands uses the widely-adopted parametric models MANO and NIMBLE, providing a more scalable and versatile solution. It also derives unified hand joints from the meshes, which facilitates seamless integration into various hand-related tasks.
Experiments on the FreiHAND and InterHand2.6M datasets demonstrate its ability to precisely reconstruct hand mesh vertices and keypoints, effectively capturing high-degree articulation motions. Empirical studies involving nine participants show a clear preference for our unified joints over existing configurations for accuracy and naturalism ($p$-value 0.016).
} 

\keywords{3D hand representation, Hand Motion Capture, Free-hand interactions, Hand dataset annotation, AR/VR}

\begin{document}


\firstsection{Introduction}

\maketitle

Capturing 3D hand motion data from real world is essential for applications such as XR hand tracking, animation, and hand synthesis. 
Common hand-capturing methods include Keypoint-skeleton tracking, which is preferred for real-time, efficient motion capture but lacks surface information, and mesh-based capturing, which provides superior detail for physical interactions but is time-intensive and costly annotations.

A unified approach is needed to integrate skeleton and mesh tracking into a standardized, high-fidelity representation. 
%
We introduce UniHands, a novel method for creating personalized hand models from wild-collected keypoints. By leveraging popular parametric hand models, MANO~\cite{romero2022embodied} and NIMBLE~\cite{li2022nimble}, UniHands reconstructs personalized hands and offers a scalable, versatile solution that integrates seamlessly with 3D modeling and XR applications. Additionally, it derives unified hand joints from the reconstructed meshes for smoother integration into various tasks.
Experiments on the FreiHAND~\cite{zimmermann2017learning} and InterHand2.6M~\cite{moon2020interhand2} datasets show that our reconstructions achieve precise alignment between keypoints and mesh vertices ($PJ/PV < 0.1 mm$) when the data format matches our optimization models. A further pilot empirical study with nine participants further evaluated the hand reconstruction subjectively when the wild-collected joint data does not match our models. Results indicated high fidelity in gesture and motion representations, with participants significantly preferring our unified joint set over existing configurations ($p$-value 0.016).


\section{Methods}
The overview of UniHands is shown in Fig.~\ref{fig:teaser} of two major components: hand mesh reconstruction from wild-collected keypoints and hand joints derivation from the reconstructed mesh.

\subsection{Mesh Reconstruction}
We designed a coarse-to-fine optimization method to refine the pose and shape ($\theta, \beta$) parameters of the MANO~\cite{romero2022embodied} hand model and its global wrist rotation ($R_w$) using gradient descent, based on input keypoints and derived joints from the reconstructed mesh.
\vspace{-1em}
\begin{equation}
    E_{key} = \sum_{i}^{N} || \mathbf{k}_i - \mathbf{J}_i(\theta, \beta, R_w) || ^2 
\end{equation}
\vspace{-1em}

where $\mathbf{k}_i$ are the input keypoints ($i \in [1,...,N]$) and $\mathbf{J}_i(\theta, \beta, R_w)$ are the derived joints from the hand mesh.

\noindent\textbf{Coarse Stage:} Start with an initial hand pose $\Bar{\theta}$ and mean shape $\Bar{\beta}$, optimizing for wrist rotation $R_w$.\\
\noindent\textbf{Fine Stage:} Use two optimizers: one to refine pose and shape parameters, and another to fine-tune $R_w$ using Adam. Besides $E_{key}$, we apply 3D mesh regularizations for integrity and smoothness.

\vspace{-1.5em}
\begin{equation}
    E_{reg} + E_{smooth} = \| \mathbf{\beta} \|^2 + \| \mathbf{\theta} \|^2 + \sum_{i} \sum_{j \in N(i)} \| \mathbf{v}_i - \mathbf{v}_j \|^2 
\end{equation}
\vspace{-1.5em}

where $\mathbf{v}_i$ and $\mathbf{v}_j$ are adjacent mesh vertices, and $N(i)$ denotes the set of neighboring vertices of vertex $i$.
The total alignment error adds up to:
\vspace{-1.5em}
\begin{equation}
E = E_{key} + \lambda_{reg} E_{reg} + \lambda_{smooth} E_{smooth}  
\end{equation}
\vspace{-2em}
 
\subsection{Unified Joints from Hand Mesh}
\begin{figure}[ht!]
\centering
 \includegraphics[width=\columnwidth]{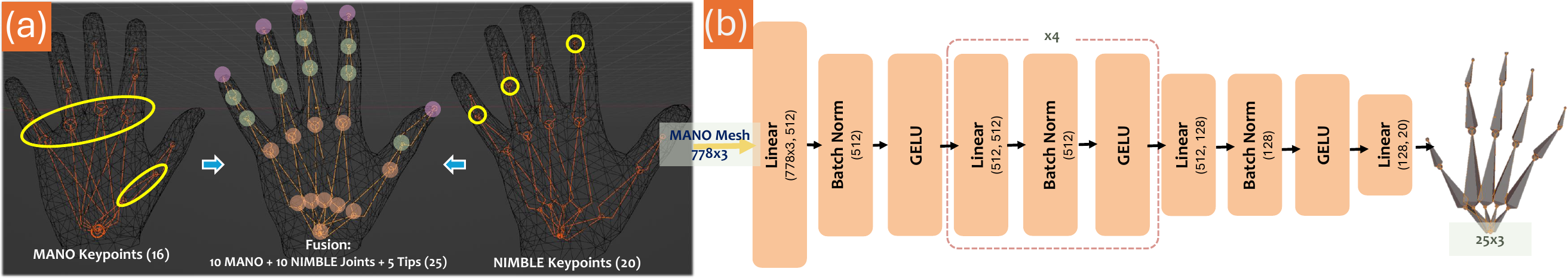}
 \vspace{-1.5em}
 \caption{\textbf{Method illustrations.} (a) Our unified joints (middle) combines the strengths of both MANO (left) and NIMBLE (right) keypoints. (b) Our MLP derives NIMBLE joints from MANO mesh.}
 \label{fig:fusion-joints}
\end{figure}

The MANO model identifies 16 hand joints but lacks complete anatomical accuracy. The NIMBLE model, developed from CT scans, offers better alignment but may misalign in new poses since it models bones and skin separately. We address this by aligning the meshes from both models and fusing their skeletons (Fig.~\ref{fig:fusion-joints} (a)).




To effectively derive joints from MANO, we first aligned the MANO mesh with NIMBLE using one-time optimizations. Then, we trained a multi-layer perceptron (MLP) to directly predict NIMBLE-like joints from the MANO mesh (Fig.~\ref{fig:fusion-joints} (b)).


\vspace{-0.5em}
\vspace{-1em}



\section{Experiments}
\subsection{Evaluation of Mesh Reconstruction}

Tab.~\ref{tab:comp} presents our reconstruction evaluations on FreiHAND~\cite{zimmermann2017learning} and InterHand2.6M~\cite{moon2020interhand2} datasets. We calculated the mean position errors per joint and vertex (\textit{PJ/PV}) and their standard deviations (\textit{PJ-std/PV-std}) between the estimated and actual vertices/keypoints. The results demonstrate high accuracy ($PJ/PV < 0.1 mm$), indicating effective mesh reconstruction when keypoints derived from wild-collected 3D hand models are aligned with ours.

\vspace{-0.5em}
\begin{table}[ht!]
\centering
\begin{tabular}{c|ccccccc}
\hline
\textbf{Dataset} & \textbf{PJ} & \textbf{PV} & \textbf{PJ-std} & \textbf{PV-std}\\
\hline
FreiHAND~\cite{zimmermann2017learning} & 0.027  & 0.022 & 1.029 & 1.77\\
InterHand2.6M~\cite{moon2020interhand2} & -0.0047 & -0.012 & 1.29  &2.46\\\hline
\end{tabular}
\vspace{0.5em}
\caption{Hand reconstruction evaluation results on FreiHAND~\cite{zimmermann2017learning} and InterHand2.6M~\cite{moon2020interhand2} datasets in millimeters (mm).}\label{tab:comp}
\end{table}
\vspace{-1.5em}

\subsection{Study of Joints from Mesh}
We conducted an empirical study to evaluate hand reconstructions when wild-collected joint data \textit{does not} align with our optimized models to validate the practical applicability of our approach.


\noindent\textbf{Participants:} Nine participants with varying expertise levels.

\noindent\textbf{Study Material:} Nine reconstructed hands were randomly selected from three datasets (Shrec21, Shrec22, and Thunder01), each collected with different technologies (Ultraleap, HoloLens2, and Mediapipe). Each hand is represented with three sets of joints: MANO, NIMBLE, and our unified model.

\noindent\textbf{Task Design:} Participants assessed the reconstructed hand meshes and derived joints in Blender scenes across three tasks:

\noindent[T1] \underline{Realism}: How realistic the derived hand joints are in the reconstructed mesh?\\
\noindent[T2] \underline{Accuracy}: How closely the joints match the raw gesture data?\\
\noindent [T3] \underline{Naturalism}: How natural the skin looks in the motions?



\begin{figure}[ht!]
 \includegraphics[width=\columnwidth]{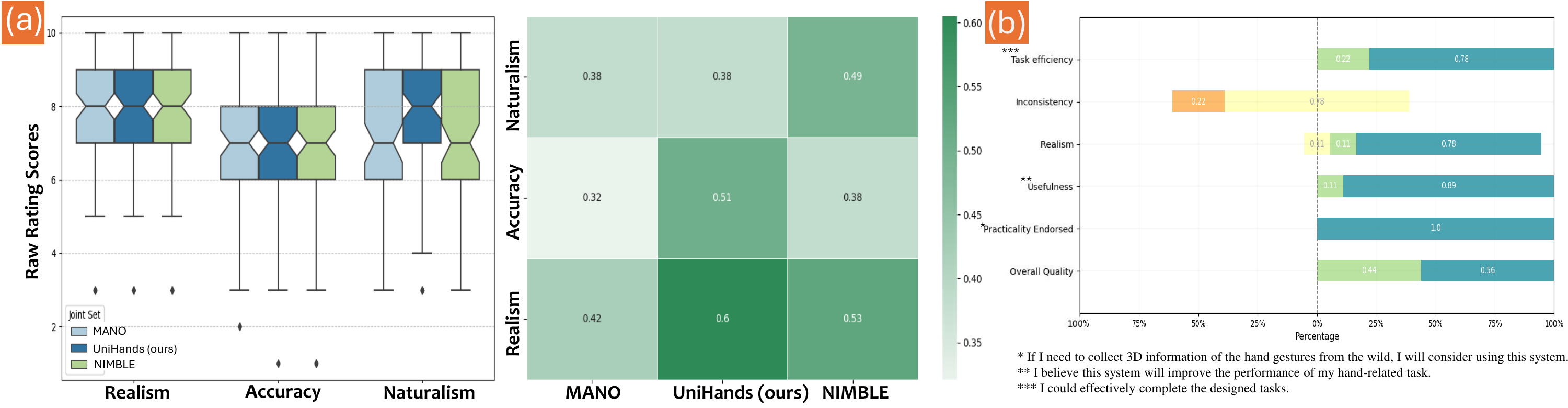}
\vspace{-1.5em}
 \caption{\textbf{Evaluation Results.} (a) Evaluations of three tasks across three joint sets, showing raw values (left) and mapped ranks (right). (b) Results from the modified System Usability Scale (SUS)/Task Load Evaluation (TLE) are weighted and displayed on a spectrum: lower ratings appear on the left in warmer hues, while higher ratings are on the right in cooler hues.}
 \label{fig:study-res}
\end{figure}

\noindent\textbf{Results:} 81 data points were collected per participant (i.e., 3 tasks $\times$ 9 gestures $\times$ 3 joint sets). Participants rated each sample on a scale from 1 to 10, then selected the top-rated sample as the best (1) and rated the others as 0. Fig.~\ref{fig:study-res}(a) presents both the raw scores and mapped ranks. Box plots show similar ratings for the first two tasks across joint sets; however, the fused model significantly outperformed the others in the third task. The mapped results further demonstrate that the fused model excels in accuracy and naturalism in animation, with a statistically significant improvement($p$-value $=0.016, < 0.05$). 

Participants completed a post-study questionnaire combining a modified System Usability Scale (SUS) and Task Load Evaluation (TLE), spanning six dimensions. Results indicate that participants consistently rated UniHands highly for performance efficiency, realism, and usefulness, demonstrating strong practical endorsement for using the system in their own work.

\vspace{-0.5em}
\vspace{-1.5em}

\section{Discussion and Conclusion}
We introduced UniHands, a novel pipeline that bridges the gap between low-fidelity hand motion tracking and high-fidelity demands in XR tasks. It offers personalized, parameterized hand models, making it an essential tool for animation, synthetic data generation, and training in hand-related tasks. Evaluations on public datasets demonstrated the high accuracy and robust performance on hand mesh reconstruction. An empirical study confirmed its superior handling of natural gestures and motions, with participants awarding high marks for realism, accuracy, and naturalism. Notably, our unified MANO-NIMBLE fusion joints were consistently rated as providing the most natural appearance in skeleton animations.

Another key finding is that people are highly sensitive to subtle anomalies in hand motion, especially when dealing with high-fidelity data. Future efforts should aim to refine techniques that balance anatomical accuracy with the fidelity of wild-collected data, to ensure high-quality hand pose reconstructions. Additionally, further studies in XR environments could personalize and improve hand modeling experiences.

\acknowledgments{
}

\bibliographystyle{abbrv-doi}

\bibliography{_ref}
\end{document}